\pdfoutput=1

\documentclass[11pt]{article}

\usepackage[final]{acl}

\usepackage{times}
\usepackage{latexsym}

\usepackage[T1]{fontenc}

\usepackage[utf8]{inputenc}

\usepackage{microtype}

\usepackage{inconsolata}

\usepackage{amssymb}
\usepackage{pifont}
\newcommand{\cmark}{\ding{51}}%
\newcommand{\xmark}{\ding{55}}%

\usepackage[utf8]{inputenc} 
\usepackage[T1]{fontenc}    
\usepackage{url}            
\usepackage{booktabs}       
\usepackage{nicefrac}       
\usepackage{microtype}      
\usepackage{hyperref} 

\usepackage{microtype}
\usepackage{multirow}
\usepackage{soul}
\usepackage{mathtools}
\usepackage{amsmath,amsfonts, amsthm}
\usepackage{algorithmic}
\usepackage[ruled,vlined]{algorithm2e}
\usepackage{graphicx}
\usepackage{float}
\usepackage{lipsum}
\usepackage{subfig}
\usepackage{textcomp}
\usepackage{placeins}
\usepackage{hhline}
\usepackage{tabularx}
\usepackage{enumitem}
\usepackage{setspace}
\usepackage{booktabs}
\usepackage{adjustbox}
\usepackage[normalem]{ulem}
\usepackage{indentfirst}
\usepackage{titlesec}
\usepackage{verbatim}
\usepackage{wrapfig}
\usepackage{lipsum} 
\usepackage{tcolorbox}

\PassOptionsToPackage{dvipsnames}{xcolor,colortbl}
\RequirePackage{xcolor,colortbl}[dvipsnames] 
\definecolor{darkcerulean}{rgb}{0.03, 0.27, 0.49}
\definecolor{burgundy}{rgb}{0.7, 0.0, 0.13}
\definecolor{majorelleblue}{rgb}{0.38, 0.31, 0.86}
\definecolor{forestgreen}{rgb}{0.13, 0.55, 0.13}
\definecolor{arylideyellow}{rgb}{0.89, 0.61, 0.06}
\hypersetup{
    colorlinks=true, 
    citecolor=darkcerulean,   
    linkcolor=burgundy,    
    urlcolor=majorelleblue     
}
\usepackage{array}
\usepackage{arydshln}
\title{GRAG: Graph Retrieval-Augmented Generation}

\author{\parbox{\linewidth}{
\centering{Yuntong Hu, Zhihan Lei, Zheng Zhang, Bo Pan, Chen Ling, Liang Zhao 
} \\
{\rm Department of Computer Science\\
   Emory University \\
   Atlanta, GA 30322, USA }\\
 \texttt{\{yuntong.hu,liang.zhao\}@emory.edu} \\
}
}

\newcounter{noteHctr} 

\begin{document}
\maketitle
\begin{abstract}
Naive Retrieval-Augmented Generation (RAG) focuses on individual documents during retrieval and, as a result, falls short in handling networked documents which are very popular in many applications such as citation graphs, social media, and knowledge graphs. To overcome this limitation, we introduce Graph Retrieval-Augmented Generation (GRAG), which tackles the fundamental challenges in retrieving textual subgraphs and integrating the joint textual and topological information into Large Language Models (LLMs) to enhance its generation.  
To enable efficient textual subgraph retrieval, we propose a novel divide-and-conquer strategy that retrieves the optimal subgraph structure in linear time.
To achieve graph context-aware generation, incorporate textual graphs into LLMs through two complementary views—the text view and the graph view—enabling LLMs to more effectively comprehend and utilize the graph context.
Extensive experiments on graph reasoning benchmarks demonstrate that in scenarios requiring multi-hop reasoning on textual graphs, our GRAG approach significantly outperforms current state-of-the-art RAG methods. 
Our datasets as well as codes of GRAG are available at \url{https://github.com/HuieL/GRAG}.
\end{abstract}

\section{Introduction}
\begin{figure*}[t!]
  \centering
  \includegraphics[width=\textwidth]{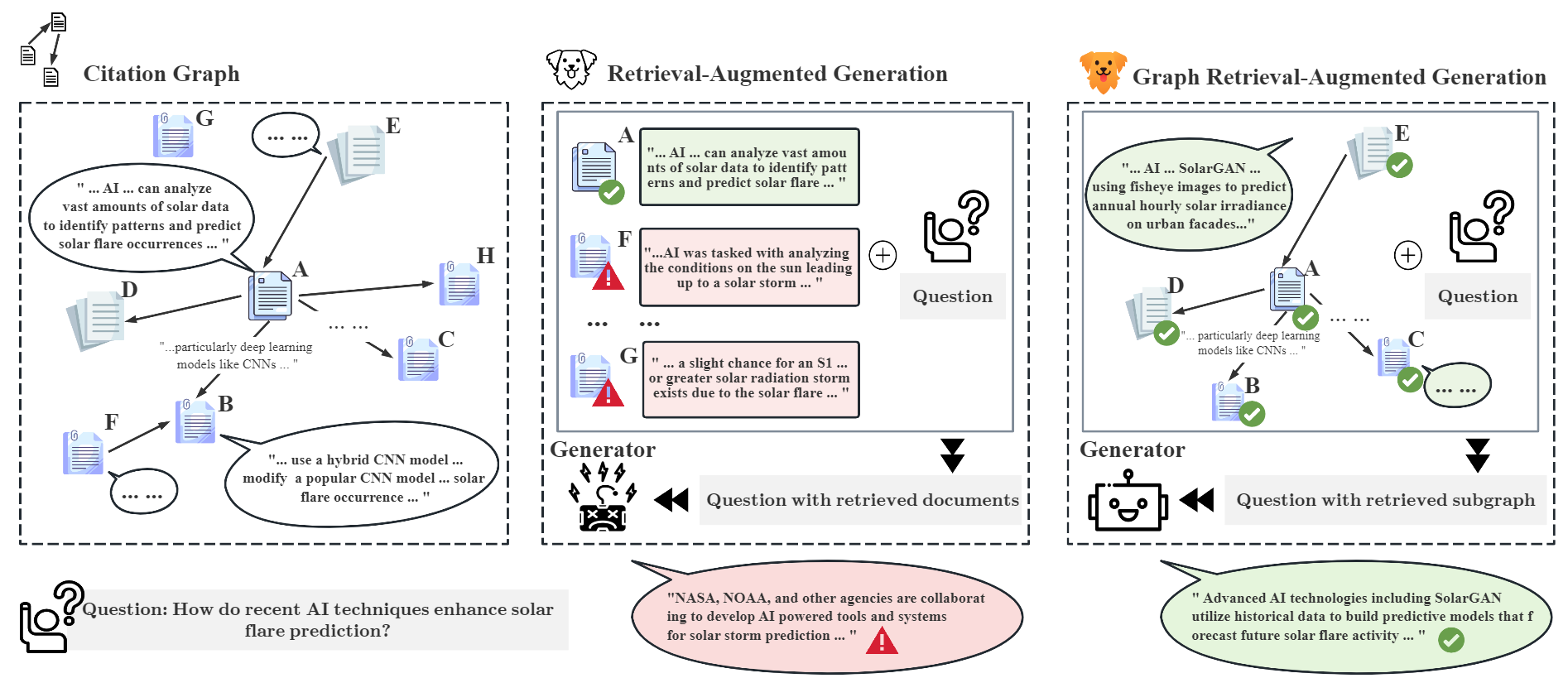}
  \caption{GRAG retrieves textual subgraphs relevant to the query, rather than discrete entities as in RAG. Entities with similar topics tend to have connections, which improves the precision and robustness of the retrieval phase.}
  \label{intro}
  \vspace{-15pt}
\end{figure*}

\noindent Large Language Models (LLMs) have demonstrated remarkable capabilities in a variety of reasoning tasks, including on graph-based data \citep{hu2023beyond, chen2024exploring, fatemi2023talk, pan2024distilling}. However, LLMs themselves struggle with factual errors due to limitations in their training data and a lack of real-time knowledge \citep{mallen2023not, min2023factscore}. Retrieval-Augmented Generation (RAG) \citep{lewis2020retrieval, guu2020realm, ram2023context, gao2023retrieval, yuspatial}, which integrates external data retrieval into the generative process, has been widely used to help LLMs access relevant information from external sources to generate more relevant answers and hence reduce factual errors \citep{tang2024multihop}. Naive RAG approaches focus solely on individual documents and retrieve relevant ones based on text similarity. However, real-world documents, such as social media postings, research papers, knowledge items, and product reviews, are typically not isolated but networked as \emph{textual graphs} \citep{he2023harnessing, jin2023large, li2023survey, hu2025cg}. Importantly, such network information is typically crucial in both retrieving relevant documents and prompting LLMs for text generation \citep{yang2024large, tang2024multihop}. For example, research papers form a citation graph, so when a solar physicist wants to learn state-of-the-art techniques in solar flare prediction, paper mutual referencing links need to be considered to pursue comprehensive retrieval coverage and insightful technical evolution understanding of this research community (as shown in \hyperref[intro]{Figure} \ref{intro}). Similarly, social interaction among social media postings, entity relations in knowledge graphs, and purchasing relations in product review systems are indispensable when LLMs want to leverage these external data. So the question is how LLMs could harness this type of networked documents when performing RAG?

To address it, we propose \textbf{Graph Retrieval-Augmented Generation (GRAG)}, which extends beyond the traditional RAG method to incorporate graph context. 
Unlike RAG, which focuses on individual documents during retrieval and generation, GRAG requires to consider the networking of documents in both stages, leading to two fundamental challenges: \textit{1) For retrieval: How to efficiently retrieve relevant textual subgraph?} Textual subgraph retrieval is particularly challenging due to the high dimensionality of textual features within nodes and edges.  \textit{2) For generation: How to deliver textual subgraph's joint textual and topological information into LLMs?}
The generation phase poses additional complexities, as it requires effectively passing networked documents to LLMs while preserving both textual and topological information, along with their interdependencies.

To address these two challenges, we propose a computational framework for GRAG. Specifically, to achieve efficient textual subgraph retrieval, we propose a new divide-and-conquer strategy that breaks the high-dimensional combinatorial optimization problem into first retrieving the most relevant ego-graphs and then refining and unioning them with a graph soft pruning mechanism.
This provides an approximate solution to identifying the most relevant textual subgraph structure, thereby avoiding the NP-hard problem of exhaustively searching all subgraphs \citep{johnson1979computers}.
To integrate the retrieved textual subgraphs into LLMs, we feed the LLM both text view via hard prompts (\textit{text tokens}) and graph view via soft prompts (\textit{graph tokens}). Retrieved textual subgraphs are transformed into hierarchical text descriptions by our proposed graph algorithms to form hard prompts, which encode the topological information in texts. Soft prompts are generated by encoding the graph's topological information directly via graph encoders, encoding text information as node/edge attributes in graphs.
Finally, the generation process is guided by both hard and soft prompts for LLM to gain a deeper understanding of the relationships between entities, leading to responses that are well-aligned with the underlying textual graph context.

Empirical results on multi-hop graph reasoning tasks demonstrate that our GRAG approach significantly outperforms RAG-based retrievers and LLM baselines in graph reasoning scenarios. In particular, Frozen LLM with GRAG outperforms fine-tuned LLM on all tasks. 

The main contributions of this article are summarized as follows:
\begin{itemize}[leftmargin=*]
\item We formulate the problem of Graph Retrieval-Augmented Generation (GRAG) and propose an efficient computational framework for GRAG, addressing the limitations of RAG methods in handling graph-based contexts.
\vspace{-2mm}
\item We propose a novel prompting method that converts textual graphs into hierarchical text descriptions without losing topological information.
\vspace{-2mm}
\item We propose an approximate solution for retrieving the most relevant textual subgraphs, efficiently avoiding the NP-hard problem of exhaustive subgraph searches. 
\vspace{-2mm}
\item Extensive experiments on graph multi-hop reasoning benchmarks demonstrate that GRAG significantly outperforms current state-of-the-art RAG methods in graph-related scenarios.
\end{itemize}

\section{Related Work}

\subsection{Prompt Tuning}
\noindent Unlike traditional fine-tuning methods, such as Low-rank Adaptation (LoRA) \citep{hu2021lora}, which require updating a model's parameters, prompt tuning focuses on modifying inputs to guide the model's responses more effectively \citep{liu2023pre, jia2022visual}. Approaches like AutoPrompt \citep{shin2020autoprompt} and Prompt Tuning \citep{lester2021power} have introduced automated techniques for crafting effective prompts without manual intervention. In particular, \citeauthor{lester2021power} propose learning soft prompts directly as embeddings, allowing task-specific adaptations while preserving the model’s original parameters. Building on this foundation, recent studies have explored adapting prompt embeddings for multi-modal contexts \citep{zhou2022learning, khattak2023maple, yang2022prompt, ge2023domain}, providing a flexible mechanism for integrating LLMs into diverse domains through prompt tuning.

\subsection{LLMs in Graph Related Tasks} 
\noindent On the one hand, the text embedding capability of LLMs helps encode textual node \& edge attributes, which directly benefits the classification task \citep{hu2023beyond, chen2023label, chen2024exploring} and knowledge graph creation \citep{trajanoska2023enhancing, yao2023exploring}. On the other hand, the contextual reasoning capabilities of the LLM benefits the graph reasoning \citep{wang2024can, jiang2023structgpt, luo2023reasoning} and graph answering in zero-shot scenarios \citep{baek2023knowledge, hu2023empirical}. 
While training on large text corpora enables LLMs to develop robust language understanding for simple graph structures, it does not inherently equip them to understand or reason about complex graph-structured data, as textual data lacks explicit topological information \citep{huang2023can, chen2024exploring, merrer2024llms}. Recently, graph prompt tuning \citep{perozzi2024let, tian2024graph} has emerged as a powerful tool to help LLMs process and comprehend topological information.


\subsection{Retrieval on Graphs}
\noindent \citeauthor{yasunaga2021qa} retrieve relevant nodes and create a joint graph that includes the QA context and the relevant nodes. \citeauthor{kang2023knowledge} and \citeauthor{kim2023factkg} focus on retrieving triples rather than individual nodes and edges to capture more complex relational data. Particularly, some retrieval problems can be solved by reasoning chains, which can be simplified to retrieve the path between the question and the target entity \citep{lo2023contextual, choudhary2023complex}. 
\citeauthor{edge2024local} leverage community detection algorithms to partition the graph into communities, then retrieve and aggregate relevant communities to generate the final answer to the query. 
\citeauthor{li2024graph} enhance retrieval processes by incorporating both textual and topological information, allowing models to better capture the structural relationships within graph-structured data.

\section{Problem Formalization}


\noindent \textbf{Textual Graphs} are graphs consisting of text-attributed nodes and edges, which can be formally defined as $G = (V, E, \{T_n\}_{n\in V}, \{T_e\}_{e\in E})$. $V$ and $E$ represent the node set and edge set. $T_n$ and $T_e$ represent the natural language attributes of the corresponding nodes and edges in the graph.

\vspace{2mm}
\noindent \textbf{Textual Subgraphs} are subgraph structures in a textual graph, e.g., $G$ with finite node set $V$ and edge set $E$, we have its subgraph set $\mathcal{S}(G) = \{g= (V', E', \{T_n\}_{n\in V'}, \{T_e\}_{e\in E'})|V'\in \mathcal{P}(V), E'\in \mathcal{P}(E)\}$, where $\mathcal{P}(V)$ and $\mathcal{P}(E)$ represent the power set of $V$ and $E$, respectively.

\vspace{2mm}
\noindent \textbf{Graph Retrieval Augmented Generation (GRAG)} aims to integrate graph context into both the retrieval and generation phases, improving the relevance of generated content to the knowledge embedded within the textual graph.
Given a specific query \( q \) over a textual graph \( G \), there exists an optimal textual subgraph \( \hat{g} \in \mathcal{S}(G) \) that leads the LLM to generate answers that align with expectations, where \( \mathcal{S}(G) \) denotes the set of all subgraphs of \( G \). The objective of GRAG is to retrieve the optimal subgraph \( \hat{g} \) and incorporate its information into an LLM$_\theta$ parameterized by $\theta$ to enhance the generation process. Formally, the probability distribution of the final output sequence \( Y \) is defined as follows:
\begin{equation}
    p_{\theta}(Y \mid [q, G]) = \prod\nolimits_{i=1}^{n} p_{\theta}(y_i \mid y_{<i}, [q, \hat{g}]),
    \label{output}
\end{equation}
where \( y_{<i} \) represents the prefix tokens, and $[q, \hat{g}]$ indicates the concatenation of the query and optimal subgraph information, respectively.

\begin{figure}[tb]
  \centering
  \includegraphics[width=0.5\textwidth]{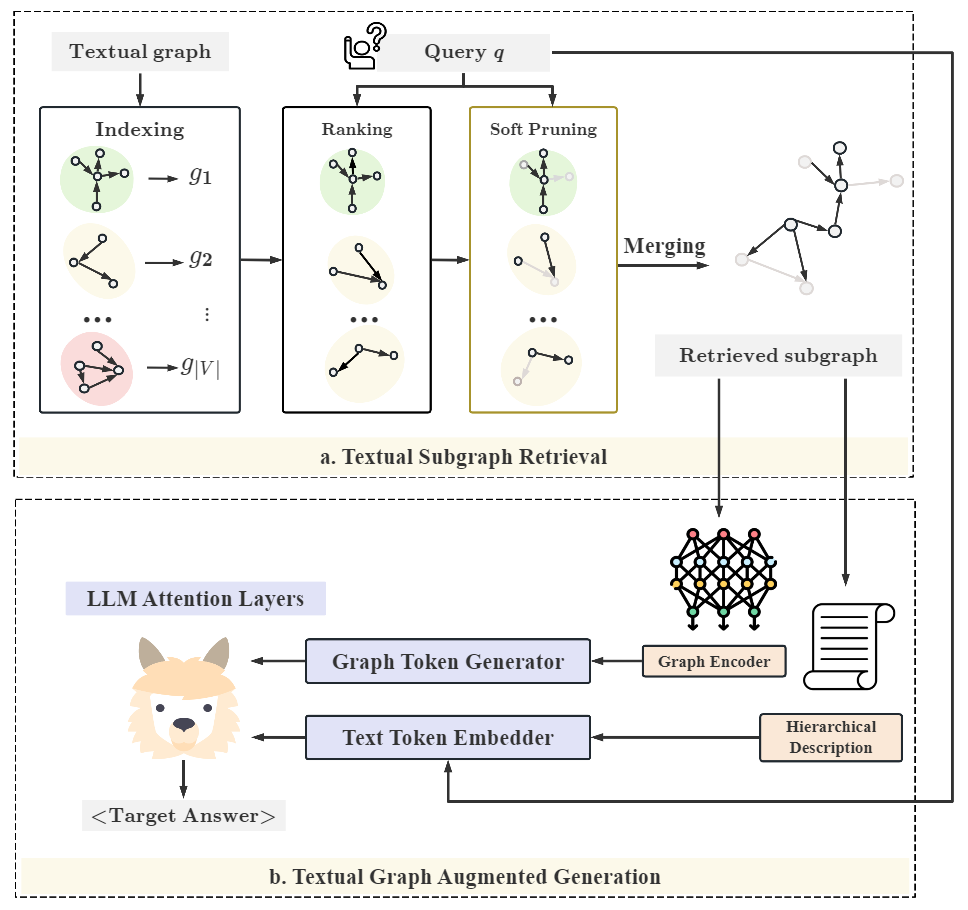}
  \captionsetup{font=footnotesize}
  \caption{Illustration of our GRAG approach.}
  \label{Flow}
  \vspace{-11pt}
\end{figure}

\section{Methodology}
\paragraph{Overview.} In this section, we introduce our solution of GRAG. As illustrated in \hyperref[Flow]{Figure 2(a)},
\textbf{to address the challenge of textual subgraph retrieval}, we propose a divide-and-conquer strategy, based on the assumption that important subgraph consists of important nodes and some of their neighbors.
specifically, we search for important ego-graphs.
We then merge the top-$N$ most relevant ego-graphs and perform soft pruning operations to reduce the impact of redundant nodes and edges, yielding an approximately optimal subgraph structure. In contrast to direct subgraph searching, which has a total search space of \(2^{|V| + |E|}\), our retrieval-then-pruning approach ensures efficiency by limiting the retrieval space to only \(|V|\) ego-graphs.
\textbf{To address the challenge of preserving both textual and topological information}, as shown in \hyperref[Flow]{Figure 2(b)}, we pursue two complementary views of textual graphs: 1) \hyperref[sec:soft]{\textit{graph view of textual graphs}}, learning representations of textual graphs as soft prompts to preserve how texts are connected, and 2) \hyperref[sec:hard]{\textit{text view of textual graphs}}, which converts the textual graph into hierarchical text descriptions as hard prompts to retain how connections are narrated. 
We then present the detailed description of the retrieval and generation processes in the following sections.

\subsection{Textual Subgraph Retrieval}

\noindent Given a textual graph $G$, the optimal textual subgraph $\hat{g} \in S(G)$ should be retrieved to maximize generation quality. Formally, let $f(\cdot)$ be the function that evaluates generation quality based on the retrieved subgraph, such that $\max_{\hat{g}}f(\hat{g})$, this problem is NP-hard and to work around it, we pursue an alternative understanding of a retrieved textual subgraph. A retrieved textual subgraph can be considered as a union of the (partial) neighborhoods of a number of important nodes, which can be formulated as:
\begin{equation}
\label{eq:dc}
    \max_{\hat{g}}f(\hat{g}) = \max_{V_{\text{key}}}f(\bigcup_{v\in{V_{\text{key}}}}G[\mathcal{N}_K^*(v)]),
\end{equation}
where $G[\mathcal{N}_K^*(v)] \in S(G)$ denotes the induced subgraph on node $v$ and its selected $K$-hop neighbors, i.e., $\mathcal{N}_K^*(v)\subseteq \mathcal{N}_K(v)$, and $V_{\text{key}}$ represents the set of key nodes that form the backbone of $\hat{g}$. 
Hence, instead of the original NP-hard problem, we approach the problem in \hyperref[eq:dc]{Equation} \ref{eq:dc} via a novel divide-and-conquer strategy that leverages the approximation: 
\begin{equation}\max_{V_{\text{key}}}f(\!\!\!\bigcup_{v\in{V_{\text{key}}}}\!\!\!G[\mathcal{N}_K^*(v)])\approx \max_{V_{\text{key}}}\!\!\!\sum_{v\in{V_{\text{key}}}} \!\!\!f(G[\mathcal{N}_K^*(v)])
\end{equation}
Hence, solving the original problem turns into selecting the top-ranked key nodes to form $V_{\text{key}}$ which has linear time complexity. More importantly, we can further accelerate it by first encoding the neighborhood surrounding each node in an offline manner. During textual subgraph retrieval, we can quickly index a pool of promising candidate subgraphs $\{G[\mathcal{N}_K(v)]\}$, from which we further rank, retain and refine the top-ranked ones.
This process is followed by a learnable pruner that carves the selected neighborhoods into subgraphs that are relevant to the query and most beneficial to the task, 
i.e., $\{G[\mathcal{N}_K(v)]\}\rightarrow \{G[\mathcal{N}_K^*(v)]\}$. 

\vspace{2mm}
\noindent \textbf{Textual Subgraph Indexing.} For any node $v$, $G[\mathcal{N}_K(v)]$ is equivalent to the $K$-hop ego-graph centered around $v$. Consequently, each $K$-hop ego-graph in $G$ is assigned a unique identifier and subsequently pooled into a graph embedding.
Specifically, we leverage a pre-trained language model (PLM)\footnote{SentenceBERT \citep{reimers2019sentence} is used to encode the query and text attributes.} to convert the text attributes of nodes and edges into embeddings. We then apply a mean pooling operation on these embeddings to obtain a graph embedding, denoted as $z_{g} \in \mathbb{R}^d$ for each subgraph $g \in \mathcal{S}(G)$, where $d$ represents the dimension of the graph embedding:
\begin{equation}
    z_{g} = \text{POOL}(\text{PLM}(\{T_n\}_{n \in V_g}, \{T_e\}_{e \in E_g})),
\end{equation}
where $V_g$ and $E_g$ represent the node set and edge set of the subgraph $g$. All indexed embeddings are stored for the subsequent retrieval process.

\vspace{2mm}
\noindent \textbf{Textual Subgraph Ranking.} The same PLM encoder is used to encode the query as:
\begin{equation}
    z_q = \text{PLM}(q) \in \mathbb{R}^d.
\end{equation}
We then calculate the semantic relevance between the query and each $K$-hop ego-graph to find the top-$N$ most relevant subgraphs:
\begin{equation}
    \mathcal{S}_N(G) = \operatorname*{Top-\textit{N}}_{g \in \mathcal{S}(G)} \text{cos}(z_q, z_g),
\end{equation}
where $\text{cos}(\cdot, \cdot)$ represents the cosine similarity function. The subset $\mathcal{S}_N(G) \subseteq \mathcal{S}(G)$ contains the $N$ subgraphs with the highest semantic relevance to the query.

\vspace{2mm}
\noindent \textbf{Textual Subgraph Soft Pruning.} Although we retrieve relevant subgraphs, some irrelevant nodes and edges may still be present, which can negatively impact the final generation. Therefore, we leverage a soft pruning approach to minimize the influence of these irrelevant entities.
Specifically, we use two Multilayer Perceptrons (MLPs) to learn a scaling factor based on the distance between nodes \& edges and the query as follows:
\begin{align}\label{coff}
  z_{n} &=\text{PLM}(T_{n}), & \alpha_{n} &= \text{MLP}_{\phi_1}~ (z_n \ominus z_q), \\
  z_{e} &=\text{PLM}(T_{e}), & \alpha_{e} &= \text{MLP}_{\phi_2}~ (z_e \ominus z_q), 
\end{align}
where $\ominus$ represents the operator to measure the element-wise distance. This scalar adaptively mask some tokens of nodes\& edges. The farther a node or edge is from the query, the closer its scalar value is to 0, effectively masking these nodes or edges. 
Finally, we merge the adaptively masked subgraphs in $\mathcal{S}_N(G)$ to obtain the optimal subgraph structure, $\hat{g}$, tailored to the query $q$, achieving this with linear complexity.

\subsection{Textual Graph Augmented Generation} 
\noindent In this section, we introduce our approach to provide LLMs with two complementary views of a textual graph: \textit{text view} and \textit{graph view}.

\vspace{2mm}
\noindent \textbf{Text View of Textual Graphs.}\label{sec:hard} 
LLMs demonstrate reasoning capabilities on graphs, particularly when interpreting texts organized in hierarchical structures, such as tree structures \citep{saad2023pdftriage}. While representing retrieved textual subgraphs in a hierarchical structure helps preserve topological information, automating this transformation remains an open challenge. Here, we propose a novel algorithm that leverages graph and tree traversals to achieve this conversion. The distinction between an ego-graph and a tree lies in the presence of additional edges connecting nodes within the same level or across multiple levels, beyond the typical parent-child connections found in a hierarchical tree structure. To overcome this challenge, 
we split each retrieved ego-graph into two parts, denoted by $g = \mathcal{T}_{g} \cup \mathcal{E}_{g}$ where $\mathcal{T}_{g}$ indicates a partially ordered set that forms a tree rooted at the ego node and $\mathcal{E}_{g}$ is an edge set consisting of edges not included in the tree. We leverage Breadth-First Search (BFS) on each ego-graph to find its $\mathcal{T}_{g}$, and then $\mathcal{E}_{g}$ can be easily obtained. Afterwards, we perform pre-order traversal on $\mathcal{T}_{g}$ and append the texts of visited node \& edge with a relation template. Then, we insert the texts of triples in $\mathcal{E}_{g}$ into the current hierarchical description.
The final description of textual graph, denoted by $D_g$, retains both textual information and topological information with a hierarchical structure, enabling lossless conversion between the $K$-hop ego-graphs and text descriptions. An example of this interconversion is provided in the \hyperref[app:graph]{Appendix} \ref{app:graph}.
Finally, we provide the LLM with a concatenation of the query and the hierarchical description of the textual subgraph (i.e., $[q, D_g]$) as a hard prompt.

\vspace{2mm}
\noindent \textbf{Graph View of Textual Graphs.}\label{sec:soft} We utilize a Graph Neural Network (GNN) to encode the graph’s topological information. To minimize the influence of irrelevant entities on generation in the encoding process, we propose to learn the representation of the soft pruned subgraph as the soft prompt. This strategy controls the message passing in the graph encoder, GNN$_{\Phi}$, through learned relevance scaling factors (\hyperref[coff]{$\alpha$}). Then, an MLP$_{\phi_3}$ is used to align the graph embeddings with the LLM tokens. This approach enables controlled message passing within GNN$_{\Phi}$, guided by the relevance between nodes \& edges and the query as follows:
\begin{align}
    m^{(l)}_u = \text{MSG}^{(l)} \left(\alpha_u \cdot h^{(l-1)}_u, \alpha_{uv} \cdot e_{uv} \right),
\end{align}
where $u \in \{ \mathcal{N}(v) \cup v \}$, $h^{(0)}_u = z_n$ and $e_{uv} = z_{uv}$, \( \mathcal{N}(v) \) represents the set of neighboring nodes of \( v \), \( h^{(l-1)}_u \) are the node features from the previous layer, \( e_{uv} \) denotes the attributes of the edge connecting nodes \( u \) and \( v \), \( \alpha_u \) and \( \alpha_{uv} \) are scaling factors. 

\vspace{2mm}
\noindent \textbf{Generation Phase.} The generation is guided by the retrieved subgraph $\hat{g}$ and the original query $q$. The modalities of these two prompts are not the same. Therefore, to bridge the gap between graph embeddings and the LLM$_{\theta}$'s text vector space, we use an MLP$_{\phi_3}$ to align the graph embeddings accordingly, as follows:
\begin{equation}
    \mathbf{h}_{\hat{g}} = \text{MLP}_{\phi_3} \left(\ \text{GNN}_{\Phi}(\hat{g}) \right)\ \in \mathbb{R}^{d_{\text{LLM}}},
\end{equation}
where \( d_{\text{LLM}} \) represents the dimension of the text vectors in LLM$_{\theta}$. \(\mathbf{h}_{\hat{g}}\) aggregates topological information to enhance LLM$_{\theta}$'s awareness of the graph's structure during the generation stage. We utilize the text embedder of LLM$_{\theta}$ to convert the hard prompt \([q, D_g]\) into text embeddings \(\mathbf{h}_T\). The final generation \( Y \) is given as follows:
\begin{align}
    & p_{\theta, \phi_1, \phi_2, \phi_3, \Phi} (Y|q, G) = p_{\theta, \phi_1, \phi_2, \phi_3, \Phi} (Y|q, \hat{g}) \nonumber\\
    & = \prod_{i=1}^{n} p_{\theta, \phi_1, \phi_2, \phi_3, \Phi} (y_i|y_{<i}, [\mathbf{h}_{\hat{g}} ; \mathbf{h}_T]),
\end{align}
where \( [\cdot;\cdot] \) denotes the concatenation of token embeddings before feeding them through transformer layers of LLM$_{\theta}$.

\section{Experiments}
\subsection{Experiment Setup}
\noindent \textbf{Datasets.} We conduct experiments on the GraphQA benchmark \citep{he2024g}. The statistics of the dataset is shown in \hyperref[datasets]{Table} \ref{datasets}. Each textual graph corresponds to at least one question-answer pair. Answering the question requires the LLM to comprehend the graph's context. {\fontfamily{pcr}\selectfont WebQSP} \citep{yih2016value, luo2023reasoning} is a large-scale, multi-hop knowledge graph QA dataset, while {\fontfamily{pcr}\selectfont ExplaGraphs} \citep{saha2021explagraphs} is a commonsense reasoning dataset focused on predicting positions in debates. 

\begin{table}[htb]
\footnotesize
\centering
\captionsetup{font=footnotesize, skip=0.10mm}
\caption{Average Dataset Statistics: the average number (\#) of graphs, nodes, edges, and tokens.}
\label{datasets}
\begin{tabular}{lcc} \toprule
\textbf{Dataset} &  {\fontfamily{pcr}\selectfont WebQSP} & {\fontfamily{pcr}\selectfont ExplaGraphs} \\ \midrule
\# Graphs & 4,700 & 2,766 \\ \hdashline
\# Nodes & 1370.89 & 5.17 \\ \hdashline
\# Edges & 4252.37 & 4.25 \\ \hdashline
\# Tokens & 100,627 & 1,396  \\ 
\bottomrule
\vspace{-6mm}
\end{tabular}
\end{table}

\vspace{2mm}
\noindent \textbf{Evaluation Metrics.} For the large-scale dataset {\fontfamily{pcr}\selectfont WebQSP}, we utilize the $F_1$ Score, Hit@1, and Recall metrics to comprehensively evaluate performance of models. For {\fontfamily{pcr}\selectfont ExplaGraphs} which focuses on common-sense reasoning, we employ Accuracy (Acc) as the primary metric. 

\vspace{2mm}
\noindent \textbf{Comparison Methods.}\label{setups} To demonstrate the effectiveness of GRAG, we compare its performance to widely used retrievers on graph multi-hop reasoning tasks .We compare GRAG with RAG using different retrievers: BM25 \citep{robertson2009probabilistic}, MiniLM-L12-v2 \citep{reimers2019sentence}, LaBSE \citep{feng2022language}, mContriever \citep{izacard2021unsupervised}, E5 \citep{wang2022text}, and G-Retriever \citep{he2024g}. Detailed introduction of comparison retrievers are presented in \hyperref[app:compare]{Appendix} \ref{app:compare}. Additionally, we establish two LLM baselines without retrieved external knowledge: (1) a frozen LLM, and (2) a fine-tuned LLM using LoRA \citep{hu2021lora}. The LLM used is the {\fontfamily{pcr}\selectfont Llama2-7b} model \citep{touvron2023llama}. Detailed experimental settings are provided in \hyperref[imp]{Appendix} \ref{imp}.

\subsection{Main Results} 
\noindent \hyperref[main_results]{Table} \ref{main_results} reports the overall results across datasets. 
We compare the performance of GRAG with comparison retrievers and baselines introduced in \hyperref[setups]{Section} \ref{setups} and make the following key observations.

\vspace{2mm}
\noindent \textbf{GRAG surpasses RAG and LLM baselines.} Notably, GRAG significantly outperforms the fine-tuned LLM in all metrics across both datasets 
without fine-tuning the LLM. Fine-tuning offers only marginal performance gains when GRAG is employed, as evidenced by the limited improvement on the {\fontfamily{pcr}\selectfont WebQSP} dataset, with the Hit@1 metric increasing from 0.7236 to 0.7275. This suggests that GRAG is a more effective strategy for enhancing the graph reasoning capabilities of LLMs than mere fine-tuning. This can significantly reduce the cost of training LLMs for graph-related tasks.
\begin{table*}[t!]\footnotesize
\centering
\vspace{-10pt}
\footnotesize
\captionsetup{font=footnotesize, skip=0.10mm}
\caption{Performance comparison across {\fontfamily{pcr}\selectfont WebQSP} and {\fontfamily{pcr}\selectfont ExplaGraphs} datasets. \textbf{Bold} numbers indicate the best performance among all models. \colorbox{majorelleblue!25}{Highlight} numbers demonstrate the performance improvement achieved by our GRAG approach compared to the LLM baselines. }
\label{main_results}
\begin{tabular}{l|c|c|ccc|c}
\hline
\textbf{\multirow{2}*{Model}} & \textbf{\multirow{2}*{Prompt tuning}} & \textbf{\multirow{2}*{Fine-tuning}} &  \multicolumn{3}{c|}{{\fontfamily{pcr}\selectfont WebQSP}} & {\fontfamily{pcr}\selectfont ExplaGraphs}\\ 
 \cline{4-7} & & &  $F_{1}$ \textbf{Score} $\uparrow$ & \textbf{Hit@1} $\uparrow$ & \textbf{Recall} $\uparrow$ & \textbf{Acc} $\uparrow$\\\hline
\multicolumn{7}{c}{{\fontfamily{pcr}\selectfont Baselines}} \\ \hline
\textbf{LLM only} & \color{burgundy}\xmark & \color{burgundy}\xmark & 0.2555 & 0.4148 & 0.2920 & 0.3394 \\
\textbf{$\text{LLM}_{LoRA}$} & \color{burgundy}\xmark & \color{forestgreen}\cmark & 0.4295 & 0.6186 & 0.4193 & 0.8927 \\\hline
\multicolumn{7}{c}{{\fontfamily{pcr}\selectfont Compared Retrievers}} \\ \hline
\textbf{BM25} & \color{burgundy}\xmark & \color{burgundy}\xmark & 0.2999 & 0.4287 & 0.2879 & 0.6011\\\hdashline
\textbf{MiniLM-L12-v2} & \color{burgundy}\xmark & \color{burgundy}\xmark & 0.3485 & 0.4730 & 0.3289 & 0.6011\\\hdashline
\textbf{LaBSE} & \color{burgundy}\xmark & \color{burgundy}\xmark &0.3280  & 0.4496 & 0.3126 & 0.6011\\\hdashline
\textbf{mContriever-Base} & \color{burgundy}\xmark & \color{burgundy}\xmark & 0.3172  & 0.4453 & 0.3047 & 0.5866 \\\hdashline
\textbf{E5-Base} & \color{burgundy}\xmark & \color{burgundy}\xmark & 0.3421 & 0.4705 & 0.3254 & 0.6011\\\hdashline
\textbf{G-Retriever} & \color{forestgreen}\cmark & \color{burgundy}\xmark & 0.4674 & 0.6808 & 0.4579 & 0.8825 \\ 
\textbf{$\text{G-Retriever}_{LoRA}$} & \color{forestgreen}\cmark & \color{forestgreen}\cmark & 0.5023 & 0.7016 & 0.5002 & 0.9042 \\ \hline
\multicolumn{7}{c}{{\fontfamily{pcr}\selectfont Our Retrieval Approach}} \\ \hline
\textbf{GRAG} & \color{forestgreen}\cmark & \color{burgundy}\xmark & 0.5022 & 0.7236 & 0.5099 & 0.9223\\
$\Delta_{LLM}$ & & & \cellcolor{majorelleblue!25}{$\uparrow$ 96.56\%} & \cellcolor{majorelleblue!25}{$\uparrow$ 74.45\%} & \cellcolor{majorelleblue!25}{$\uparrow$ 74.62\%} & \cellcolor{majorelleblue!25}{$\uparrow$ 171.74\%}\\ \hdashline
\textbf{$\text{GRAG}_{LoRA}$} & \color{forestgreen}\cmark & \color{forestgreen}\cmark & \textbf{0.5041} & \textbf{0.7275} & \textbf{0.5112} & \textbf{0.9274}\\
$\Delta_{LoRA}$ & & & \cellcolor{majorelleblue!25}{$\uparrow$ 17.37\%} & \cellcolor{majorelleblue!25}{$\uparrow$ 17.60\%} & \cellcolor{majorelleblue!25}{$\uparrow$ 21.92\%} & \cellcolor{majorelleblue!25}{$\uparrow$ 3.89\%}\\ \hline
\end{tabular}
\vspace{-10pt}
\end{table*}

\vspace{2mm}
\noindent \textbf{Soft pruning boosts LLM performance in graph-related tasks.} When all textual information from graphs is integrated into the prompt, the LLM exhibits suboptimal performance, even on the {\fontfamily{pcr}\selectfont ExplaGraphs} dataset, which features smaller graph sizes. This underscores the critical need to implement retrieval operations to mitigate the negative impact of redundant information in graphs. Notably, fine-tuning yields significant improvements in the performance of the LLM when reasoning on small graphs, with a notable increase from 33.94\% to 89.27\% accuracy on {\fontfamily{pcr}\selectfont ExplaGraphs}. However, the benefits of fine-tuning diminish with larger graph sizes, with Hit@1 on {\fontfamily{pcr}\selectfont WebQSP} only increasing from 0.4148 to 0.6186.

\vspace{2mm}
\noindent \textbf{GRAG demonstrates strong transferability} to transfer learned textual graph encoding capabilities across datasets. As shown in \hyperref[across]{Table} \ref{across}, when trained on a large dataset, GRAG can enhance generation on a smaller dataset using the trained model. Notably, GRAG trained on {\fontfamily{pcr}\selectfont WebQSP} on {\fontfamily{pcr}\selectfont ExplaGraphs} outperforms the naive LLM, with an accuracy improvement of 33.77\%.

\begin{table}[htb]
\footnotesize
\centering
\captionsetup{font=footnotesize, skip=0.10mm}
\caption{Cross-Dataset Transfer Learning Performance.}
\makeatletter\def\@captype{table}
\begin{tabular}{lcc} \toprule
\textbf{Transferability} & \textbf{Acc} & $\Delta_{LLM}$ \\ \midrule
{\fontfamily{pcr}\selectfont WebQSP} $\rightarrow$ {\fontfamily{pcr}\selectfont ExplaGraphs} & 0.4540 & \cellcolor{majorelleblue!25}{$\uparrow$ 33.77\%}\\ \midrule
 & \textbf{Hit@1} & $\Delta_{LLM}$ \\ \midrule
{\fontfamily{pcr}\selectfont ExplaGraphs} $\rightarrow${\fontfamily{pcr}\selectfont WebQSP} & 0.4237 & \cellcolor{majorelleblue!25}{$\uparrow$ 2.15\%}\\ 
\bottomrule
\end{tabular}
\label{across}
\vspace{-13pt}
\end{table}

\vspace{2mm}
\noindent \textbf{Larger LLMs don’t necessarily outperform smaller ones in graph-related tasks without retrieval.} Beyond the performance comparison of GRAG and RAG models, we evaluated the impact of LLM scale on graph-related tasks, specifically examining the 7B and 13B versions of the Llama model. Our findings indicate that larger LLMs may underperform relative to smaller models.
In the absence of retrieval techniques, larger LLMs fail to yield superior performance in these tasks. For example, the {\fontfamily{pcr}\selectfont llama2-7b-chat-hf} model achieves an accuracy of 33.94\% on the commonsense reasoning task in the {\fontfamily{pcr}\selectfont ExplaGraphs} dataset, marginally outperforming the {\fontfamily{pcr}\selectfont llama2-13b-chat-hf} model, which records an accuracy of 33.57\%. A similar trend is observed on the {\fontfamily{pcr}\selectfont WebQSP} dataset, where the 13B model's Hit@1 score of 0.4112 is slightly lower than the 0.4148 achieved by the 7B model.

\begin{figure}[bh]
  \centering
  \vspace{-2mm}
  \includegraphics[width=0.36\textwidth]{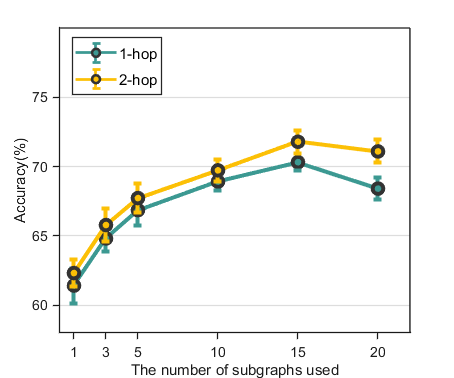}
  \captionsetup{font=footnotesize}
  \caption{Performance of our GRAG approach on {\fontfamily{pcr}\selectfont WebQSP} as the ego-graph size and number of ego-graphs used vary.}
  \label{tradeoff}
  \vspace{-6mm}
\end{figure}

\subsection{Discussion}
\noindent \textbf{Impact of Subgraph Size $K$}. The retrieval efficiency of our GRAG approach is preserved by constraining the search space to only $|V|$ $K$-hop ego-graphs. However, as the subgraph size $K$ increases, a broader range of graph context is integrated during the generation process, resulting in longer training and inference times. Additionally, embeddings of larger subgraphs (i.e., over 3-hop) are more susceptible to oversmoothing, which can diminish their distinctiveness for retrieval. Therefore, the subgraph size must be carefully controlled to avoid excessive growth. \hyperref[tradeoff]{Figure} \ref{tradeoff} shows the performance of GRAG on {\fontfamily{pcr}\selectfont WebQSP} as the number of $1$-hop and $2$-hop ego-graphs changes. With the same number of retrieved ego-graphs, using $2$-hop ego-graphs consistently outperform using $1$-hop ego-graphs. 
Increasing the number of retrieved subgraphs does not necessarily improve generation quality due to the introduction of more irrelevant information. A drop in performance is observed when the number of ego-graphs increases from 15 to 20. Moreover, using a larger number of subgraphs results in more robust generation, as indicated by a smaller standard deviation. 

\vspace{2mm}
\noindent \textbf{Evaluation on Hallucinations.} We conduct a small-scale human evaluation of GRAG outputs to assess hallucinations. Specifically, we randomly select and manually review 100 samples from the {\fontfamily{pcr}\selectfont WebQSP} and {\fontfamily{pcr}\selectfont ExplaGraphs} results. The LLM is prompted to generate answers to the questions, along with the referenced nodes and edges. Following \citeauthor{menick2022teaching} and \citeauthor{he2024g}, human annotators evaluate whether the model output is reasonable and supported, verifying whether nodes and edges referenced in the output exist in the actual graph. GRAG’s outputs reference 79\% of valid entities in the graph, compared to MiniLM-L12-v2 and G-Retriever, which reference 62\% and 71\% of valid entities, respectively.

\vspace{2mm}
\noindent \textbf{Thorough Comparisons with RAG.}
Overall, the LLM can generate better responses with retrieved entities by all tested retrievers. However, even if advanced retrievers use more data for training and increase the embedding dimension to obtain better embeddings, their focus remains exclusively on the text domain, creating a performance bottleneck as no topological information is retrieved. As shown in \hyperref[main_results]{Table} \ref{main_results}, when graph context is not considered, there is only a slight difference in the enhancement achieved by various retrievers. This phenomenon is further discussed in \hyperref[appexp]{Appendix} \ref{appexp}. G-Retriever, which aggregates topological information as soft prompts, outperforms other retrievers, but it also fails to consider topology during the retrieval process. Our GRAG approach addresses this limitation by directly retrieving subgraphs instead of individual entities and incorporating topological information into the LLM during the generation phase, thereby achieving optimal performance on both datasets.

\vspace{-2mm}
\subsection{Ablation Study}
\noindent We conducted a series of ablations to our GRAG framework to identify which components play a key role. We evaluate four model variants trained differently, where fine-tuning is used and 2-hop ego-graphs are retrieved in all settings: \textit{w/o Retrieval} trains the LLM without retrieving subgraphs, instead providing the entire graph to the LLM. \textit{w/o Graph Encoder} trains the LLM using the text on the retrieved textual subgraphs, but does not generate graph tokens to provide the graph context; \textit{w/o Soft Pruning} indicates that irrelevant entities are not pruned when retrieved subgraphs are encoded to the graph tokens; \textit{w/o Graph Description} trains the LLM without the hierarchical text descriptions of retrieved textual subgraphs. \hyperref[ablation]{Table} \ref{ablation} shows the main results.
Our main findings are as follows:

\vspace{2mm}
\noindent \textbf{Importance of Graph Context.} When the graph context is not encoded (\textit{w/o Graph Encoder}), the LLM’s generation quality significantly declines (Hit@1: 0.7275 $\rightarrow$ 0.5835). This suggests that merely describing relationships between nodes and edges in text is insufficient for LLMs to fully comprehend the graph context. Embedding the graph enables the LLM to capture the graph’s context at a deeper level.

\begin{table}[thb]
\footnotesize
\centering
\captionsetup{font=footnotesize}
\caption{Ablation study on {\fontfamily{pcr}\selectfont WebQSP}.  $\Delta_{GRAG}$ represents the change in Hit@1 performance relative to our full GRAG approach.}
\vspace{-2mm}
\makeatletter\def\@captype{table}
\begin{tabular}{lcc} \toprule
\textbf{Setting} &  \textbf{Hit@1} & $\Delta_{GRAG}$ \\ \midrule
w/o Retrieval & 0.6093 & \cellcolor{majorelleblue!25}{$\downarrow$ 16.25\%} \\ \hdashline
w/o Graph Encoder & 0.5835 & \cellcolor{majorelleblue!25}{$\downarrow$ 19.79\%} \\ \hdashline
w/o Soft Pruning & 0.5671 & \cellcolor{majorelleblue!25}{$\downarrow$ 22.05\%} \\ \hdashline
w/o Graph Descriptions  & 0.4496 &  \cellcolor{majorelleblue!25}{$\downarrow$ 38.20\%}  \\ 
\bottomrule
\end{tabular}
\label{ablation}
\vspace{-5mm}
\end{table}

\vspace{2mm}
\noindent \textbf{Impact of Pruning.} When irrelevant entities in retrieved textual subgraphs are not pruned (\textit{w/o Soft Pruning}), the performance on {\fontfamily{pcr}\selectfont WebQSP} is worse compared to the \textit{w/o Retrieval} and \textit{w/o Graph Encoder} variant. This suggests that pruning is crucial, especially in dense graphs, to improve the quality of graph tokens and avoid negative impacts from irrelevant entities.

\vspace{2mm}
\noindent \textbf{Importance of Text Attributes.} When the text attributes of retrieved subgraphs are excluded, relying solely on the soft token does not enhance the generation process in graph-related tasks. This variant performs worse than the \textit{w/o Retrieval} setup, with its Hit@1 score dropping to 0.4496—a 38.2\% decrease. This finding highlights the importance of node and edge textual attributes for effective generation. While soft tokens aggregate these text attributes, incorporating the text attributes remains essential for optimal LLM generation.

\section{Conclusion}
\noindent In this paper, we introduce Graph Retrieval-Augmented Generation (GRAG) to extend Retrieval-Augmented Generation (RAG) to graph-based scenarios. We present a computational framework for GRAG that enhances the generation capabilities of Large Language Models (LLMs) by retrieving query-relevant textual subgraphs. To ensure efficient subgraph retrieval, we propose a divide-and-conquer strategy that leverages $K$-hop ego-graphs and soft pruning to approximate the optimal textual subgraph. Our approach provides LLMs with two complementary views of a textual graph: \textit{graph view} and \textit{text view}, enabling a comprehensive understanding of the graph context.
Empirical results demonstrate that GRAG significantly outperforms LLM baselines and RAG-based LLMs, particularly in scenarios requiring detailed, multi-hop reasoning on textual graphs. Our approach not only addresses the NP-hard challenge of exhaustive subgraph searches but also shows that a frozen LLM enhanced by GRAG can outperform fine-tuned LLMs at a reduced training cost.

\section{Limitations}
\noindent While GRAG provides an effective framework for graph context-aware generation, the efficiency of GRAG's textual subgraph retrieval depends on the quality of the initial node ranking and pruning mechanism. In cases where graph structure or node importance is difficult to estimate, this could lead to suboptimal retrieval performance.

\section*{Acknowledgments}
This work was supported by the National Science Foundation (NSF) Grant No. 2414115, No. 2403312, No. 2007716, No. 2007976, No. 1942594, No. 1907805, NIH R01AG089806, and NIH R01CA297856.

\bibliography{custom}

\appendix
\section{Appendix}
\subsection{Hierarchical Description}\label{app:graph}
\begin{figure}[h]
  \centering
  \includegraphics[width=0.42\textwidth]{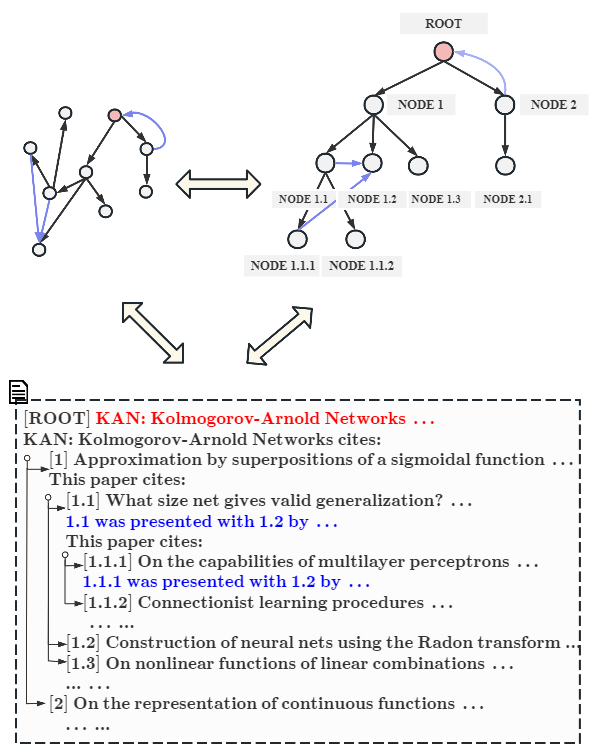}
  \caption{An Example hierarchical description for a 2-hop ego-graph from a citation network.}
  \label{tree}
\end{figure}

\noindent As shown in \hyperref[webqsp]{Figure} \ref{tree}, a 2-gop ego-graph is transformed into a nested, indented list, which mirrors the graph’s structure. Each level in the hierarchy corresponds to a level in the graph, representing connections between nodes. For example, "NODE 1" contains sub-nodes (NODE 1.1, NODE 1.2, etc.), which are further divided into lower levels, reflecting the original graph's branching structure.

\begin{table*}[h!]\small
\centering
\vspace{-10pt}
\captionsetup{font=footnotesize, skip=0.10mm}
\caption{Performance of RAG-based retrievers: Hit@1 on {{\fontfamily{pcr}\selectfont WebQSP}} and Acc on {{\fontfamily{pcr}\selectfont ExplaGraphs}}.}
\label{tab:math_results}
\begin{tabular}{l|ccccc|ccc}
\hline
\textbf{\multirow{2}*{Model}} &  \multicolumn{5}{c|}{{\fontfamily{pcr}\selectfont WebQSP}} & \multicolumn{3}{c}{{\fontfamily{pcr}\selectfont ExplaGraphs}}\\ 
 \cline{2-9} & top-3 & top-5 & top-10 & top-15 & top-20 & top-3 & top-5 & top-10 \\\hline
\textbf{BM25} & 0.3722 & 0.3821 & 0.4109 & 0.4165 & 0.4287 & 0.5704& 0.5921 & 0.6011\\\hdashline
\textbf{MiniLM-L12-v2} & 0.4251 & 0.4251 & 0.4539 & 0.4625 & 0.4730 & 0.5848 & 0.5939 & 0.6011\\\hdashline
\textbf{LaBSE} & 0.4091 & 0.4171 & 0.4294 & 0.4527 & 0.4496 &  0.6011 & 0.6011 & 0.6011\\\hdashline
\textbf{mContriever-Base} & 0.4183 & 0.4158 & 0.4349 & 0.4459 & 0.4453 & 0.5866  & 0.5866 & 0.5866 \\\hdashline
\textbf{E5-Base} & 0.4404 & 0.4558 & 0.4662 & 0.4650 & 0.4705 & 0.5921 & 0.5939 & 0.6011\\\hline
\end{tabular}
\end{table*}
Each node in the hierarchy is accompanied by descriptive text or titles, conveying the content or subject matter associated with that node. This setup preserves the textual information within the graph, such as titles of cited papers or key phrases. The hierarchical formatting maintains the topological relationships between nodes, with indentation and nested levels reflecting their connections. This structure mirrors the links between the root node and its sub-nodes, capturing the connectivity of the original graph. By presenting the graph as a sequential, hierarchical text format, the order and relationships among nodes become clear, as each node's connection to its parent and child nodes is preserved through indentation and citation references.

\subsection{Comparison Retrievers}\label{app:compare}
\noindent BM25 \citep{robertson2009probabilistic}, which is a statistical model, scores documents based on term frequency, inverse document frequency, and document length, using probabilistic principles to estimate the relevance of documents to a query; MiniLM-L12-v2, which is a SentenceTransformer model \citep{reimers2019sentence} widely used in clustering and semantic search; LaBSE \citep{feng2022language}, a BERT-based model that performs retrieval by using a dual-encoder framework to learn cross-lingual sentence embeddings; mContriever \citep{izacard2021unsupervised}, which utilizes a contrastive learning approach with a bi-encoder architecture to independently encode documents and queries; E5 \citep{wang2022text}, that employs a contrastive pre-training strategy using a bi-encoder architecture, optimizing similarity between relevant pairs while distinguishing from irrelevant ones using in-batch negatives; G-Retriever \citep{he2024g}, which retrieves relevant nodes and edges, and then constructs a relevant subgraph using a Prize-Collecting Steiner Tree method.

\subsection{Implementation}\label{imp}
\noindent The data splits for training, validation, and test sets are 60\%/20\%/20\% for {\fontfamily{pcr}\selectfont ExplaGraphs} and 60\%/5\%/35\% for {\fontfamily{pcr}\selectfont WebQSP}.
All experiments are performed on a Linux-based server with 4 NVIDIA A10G GPUs. We use SentenceBert \citep{reimers2019sentence} to encode the question and text attributes to obtain vectors for the retrieval process. The graph encoder, i.e. GAT \citep{velivckovic2018graph}, has 4 layers with 4 heads per layer and a hidden dimension size of 1024. 

The LLM backbone is {\fontfamily{pcr}\selectfont Llama-2-7b-hf}, while the model used is in the setting of LLM only is {\fontfamily{pcr}\selectfont Llama-2-7b-chat-hf}. We employ Low-rank Adaptation (LoRA) \citep{hu2021lora} for fine-tuning, configuring the LoRA parameters as follows: the dimension of the low-rank matrices is set to 8; the scaling factor is 16; and the dropout rate is 0.05. For the optimization, AdamW optimizer \citep{loshchilov2018decoupled} is used. The initial learning rate is set to 1e-5 and the weight decay is 0.05. Each experiment runs for up to 10 epochs, and the batch size is 2. For compared retrievers, each experiment on {\fontfamily{pcr}\selectfont ExplaGraphs} is replicated three times, utilizing different retrieval settings for each run, i.e., top-3, top-5 and top-10; Each experiment on {\fontfamily{pcr}\selectfont WebQSP} is replicated five times, utilizing different retrieval settings for each run, i.e., top-3, top-5, top-10, top-15 and top-20, where top-\textit{k} denotes that the \textit{k} most relevant nodes and \textit{k} edges are retrieved and used for generation. In our GRAG approach, since the graphs in {\fontfamily{pcr}\selectfont ExplaGraphs} are constructed from several triples, each graph is actually a chain consisting of only a few nodes. Therefore, we feed the entire graph to the LLM.


\subsection{Experiment} \label{appexp}
\noindent \textbf{Evaluation Metrics.} \textbf{Hit@1} assesses whether the top retrieved result is correct. It is particularly useful for understanding the accuracy of the first retrieval hit in graph-based question answering tasks. \textbf{$F_1$ Score} is the harmonic mean of precision and recall, providing a single metric that balances both false positives and false negatives. \textbf{Recall} measures the proportion of relevant entities that are successfully retrieved. High recall indicates that the retrieval system captures most of the relevant information. \textbf{Accuracy (Acc)} measures the proportion of correctly answered questions. It is particularly useful for tasks like {\fontfamily{pcr}\selectfont ExplaGraphs}, where the focus is on commonsense reasoning.
\vspace{-6pt}

\begin{figure}[h]
  \centering
  \includegraphics[width=0.42\textwidth]{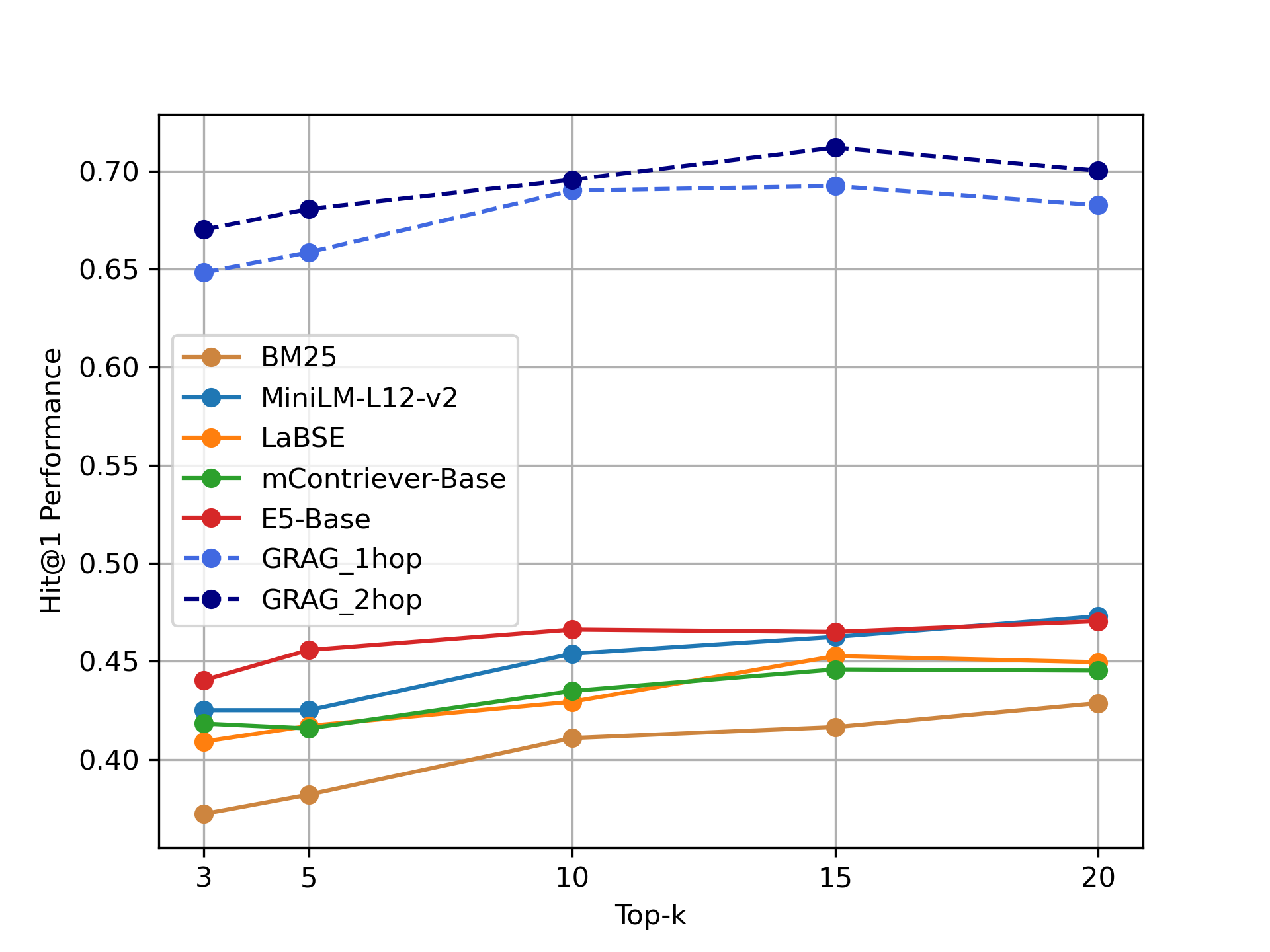}
  \caption{Effects of the number of retrieved entities on the {\fontfamily{pcr}\selectfont WebQSP} dataset.}
  \label{webqsp}
\end{figure}
\vspace{-6pt}

\noindent \textbf{Effects of the Number of Retrieved Entities.} Top-$k$ indicates $k$ nodes and $k$ edges are retrieved. The performance of various RAG retrievers on the {\fontfamily{pcr}\selectfont WebQSP} and {\fontfamily{pcr}\selectfont ExplaGraphs} datasets, with different numbers of retrieved entities, is summarized in \hyperref[tab:math_results]{Table} \ref{tab:math_results}. As shown in \hyperref[webqsp]{Figure} \ref{webqsp}, GRAG replaces hard prompts with texts of retrieved entities, while the soft prompt is represented by tokens generated from the retrieved $K$-hop ego-graphs. Increasing the number of retrieved entities generally improves performance up to a certain point. For example, BM25’s Hit@1 score on {\fontfamily{pcr}\selectfont WebQSP} rises from 0.3722 with top-3 retrievals to 0.4287 with top-20 retrievals, and MiniLM-L12-v2 shows improvement from 0.4251 to 0.4730 over the same range. However, this trend does not continue indefinitely; for some models, performance plateaus or even slightly decreases beyond a certain number of entities. For instance, LaBSE’s performance peaks at top-15 and then slightly declines at top-20 on {\fontfamily{pcr}\selectfont WebQSP}. This suggests that retrieving too many entities can introduce irrelevant information, potentially impairing final generation quality. On the {\fontfamily{pcr}\selectfont ExplaGraphs} dataset, the trend is less pronounced due to smaller graph sizes, with most models showing minimal performance changes beyond top-5 retrievals. When the graph size is small, indicating limited information, all RAG-based retrievers encounter a performance bottleneck. In contrast, our GRAG approach leverages topological information effectively, enabling it to overcome this limitation.

\end{document}